# Ecological Data Analysis Based on Machine Learning Algorithms

**Md.Siraj-Ud-Doulah[1*], and Md. Ashad Alam[2]**

[1]Department of Statistics, Begum Rokeya University, Rangpur 5400, Bangladesh**;** sdoulah_brur@yahoo.com

[2]Department of Statistics, Hajee Mohammad Danesh Science and Technology University
Dinajpur 5200, Bangladesh; ashad.hstu@gmail.com
**\*** Correspondence: sdoulah_brur@yahoo.com

**Abstract:** Classification is an important supervised machine learning method, which is necessary and challenging issue for ecological research. It offers a way to classify a dataset into subsets that share common patterns. Notably, there are many classification algorithms to choose from, each making certain assumptions about the data and about how classification should be formed. In this paper, we applied eight machine learning classification algorithms such as Decision Trees, Random Forest, Artificial Neural Network, Support Vector Machine, Linear Discriminant Analysis, k-nearest neighbors, Logistic Regression and Naive Bayes on ecological data. The goal of this study is to compare different machine learning classification algorithms in ecological dataset. In this analysis we have checked the accuracy test among the algorithms. In our study we conclude that Linear Discriminant Analysis and k-nearest neighbors are the best methods among all other methods**.**

## 1. Introduction

Over the past couple decades, machine learning has become one of the strengths of information technology and with that, a rather fundamental, notwithstanding usually hidden, part of our life described in [1, 9]. With the ever growing volumes of data becoming available there is good aim to trust that smart data analysis will become even more persistent as a essential component for technological advancement. Machine Learning aims to produce classifying languages simple enough to be understood easily by the human discussed in [2,3]. They must mimic human reasoning adequately to deliver vision into the result development. Like statistical methods, contextual information may be exploited in progress, but operation is assumed without human involvement defined in [7, 10]. Classification has two distinct meanings. We may be given a set of observations with the aim of forming the actuality of classes or clusters in the data. Or we may know for firm that there are so many classes, and the aim is to form a rule whereby we can categorize a new observation into one of the existing classes labelled in [ 24]. The former type is known as Unsupervised Learning (or Clustering), the latter as Supervised Learning. In this paper when we use the term classification, we are talking of Supervised Learning. In the statistical literature of [20], Supervised Learning is frequently, but not continuously, mentioned to as discrimination, by which is predestined the forming of the classification rule from given appropriately classified data. Classification is a data mining function that allocates items in a gathering to target groups or classes. The goal of classification is to exactly forecast the target class for each case in the data. In the model build (training) process, a classification algorithm discoveries associations between the values of the predictors and the values of the target.

Different classification algorithms practice different procedures for discovery associations. These associations are abridged in a model, which can then be functional to a different data set in which the class projects are unidentified. Classification models are verified by matching the projected values to recognize target values in a set of test data. The past data for a classification task is normally divided into two data sets: one for building the model; the other for testing the model in [14]. Recording a classification model outcomes in class projects and probabilistic for each case. Classification has many applications in purchaser separation, business forming, marketing, credit analysis, biomedical and drug response modeling and ecological dataset etc. The process of applying supervised ML to a real-world problem is described in Figure 1.

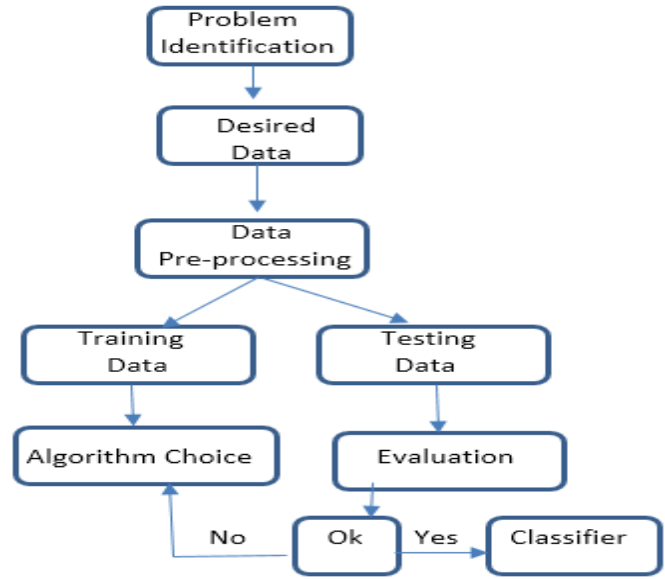

This work efforts on the classification of ML algorithms and determining the most competent algorithm with highest accuracy and precision. As well as establishing the performance of different algorithms on large and smaller data sets with a understanding classify them appropriately and provide insight on how to build supervised machine learning models. The remaining part of this work is arranged as follows: Section 2 presents the materials and methods of different supervised Machine Learning classification algorithms; also presents the performance evaluation criteria for algorithm, and discusses Software that we used here, Section 3 gives the results of the work while Section 4 gives the conclusion.

**2. Materials and Methods**

According to [18, 25, 28], the supervised machine learning algorithms which deals more with classification includes the following algorithms: Decision Trees, Random Forest, Neural Network, Support Vector Machine, Linear Discriminant Analysis, k-nearest neighbors, Logistic Regression and Naive Bayes and their performance will be evaluated.

*2.1 Algorithms*
2.1.1 Decision Tree

Decision trees are frequently used in operations research, explicitly in decision analysis, to support recognize a policy most likely to grasp a goal, but are also a common tool in machine learning. To build a decision tree which is designed by [5, 17, 26], we need to compute two types of entropy using frequency tables as follows:

a) Entropy using the frequency table of one attribute: $E(S) = \sum_{i=1}^{c} -p_i \log_2 p_i$

b) Entropy using the frequency table of two attributes: $E(T, X) = \sum_{c \in X} p(c) E(c)$

The information gain is based on the decline in entropy after a dataset is divided on an attribute. Building a decision tree is all about discovery attribute that returns the uppermost information gain

1. Compute entropy of the target.
2. The dataset is then divided on the dissimilar attributes. The entropy for every outlet is calculated. Then it is added proportionally, to get total entropy for the divided. The resulting entropy is subtracted from the entropy before the divided.
3. Select attribute with the major information gain as the decision node, split the dataset by its subdivisions and repeat the same process on each outlet.
4. An outlet with entropy of 0 is a leaf node. Next an outlet with entropy more than 0 needs further splitting.
5. The algorithm is run recursively on the non-leaf outlets, until all data is classified.

### 2.1.2. Random Forest

Random forests or random decision forests are a collaborative learning process for classification, regression and other tasks, that function by building a assembly of decision trees at training time and outputting the class that is the mode of the classes (classification) or mean forecast (regression) of the distinct trees. Random decision forests accurate for decision trees' habit of over fitting to their training set. The algorithm of random forest have been proposed by [5, 19].

1. Choice random samples from a assumed dataset.
2. Build a decision tree for each sample and catch a forecast outcome from each decision tree.
3. Complete a vote for each forecast outcome.
4. Choice the estimate result with the most votes as the finishing prediction.

### 2.1.3. Artificial Neural Network

Artificial neural networks (ANNs) are calculating systems imprecisely stimulated by the biological neural networks that establish animal brains. The algorithm of ANN have been suggested by [5, 15]. The yield of the model is figured using the following mathematical expression:

$$y_t = \alpha_0 + \sum_{j=1}^{q} \alpha_j g\left(\beta_{0j} + \sum_{i=1}^{p} \beta_{ij} y_{t-i}\right)$$

The integer's $p$, $q$ are the number of input and hidden nodes respectively. $\alpha_j (j = 0, 1, \cdots, q)$ and $\beta_{ij} (i = 0, 1, \cdots, p; j = 0, 1, \cdots, q)$ are the connection weights and $\alpha_0$ are $\beta_{0j}$ the bias terms. Usually, the logistic sigmoid function $g(x) = 1 / (1 + e^{-x})$ is applied as the nonlinear activation function. Other activation functions, such as linear, hyperbolic tangent, Gaussian, etc. can also be used.

### 2.1.4. Support Vector Machine

In machine learning, support vector machines are supervised learning models with related learning algorithms that investigate data used for classification and regression analysis. SVM is a very effective process of machine learning [27], which is based on mapping of learning cases from input space to a novel high dimensional, potentially vast dimensional feature space in which cases are linearly separable. The method then finds an optimal hyper plane [5, 11].

$$\langle w, \phi(x) \rangle + b = 0$$

Where **w** is a matrix of coefficients $\phi(x)$ is a mapping function, and $b$ is a constant. This hyper surface splits learning examples with a maximal margin. Support vectors are a small set of critical border examples of each class, best separated by this hyper plane. Building of an optimal hyper plane is performed using iterative algorithm which minimizes the error estimation function:

$$\frac{1}{2} w'w + c \sum_{i=1}^{n} \xi_i$$

with the constraints $y_i \left( w'\phi(x_i) + b \right) \geq 1 - \xi_i$, $i = 1, 2, \cdots, N$, $\xi \geq 0$, $i = 1, 2, \cdots, n$

where $w$ is a vector of coefficients, $b$ is a constant, $\xi$ is a slack variable, $n$ is a number of learning examples and $C$ is a regularization parameter. SVM method uses linear functions to create discrimination borders in a high dimensional space. Non-linear discriminant function in an input space is obtained using inverse transformation.

### 2.1.5. Linear Discrimination Analysis

In the case where there are more than two classes, the analysis used in the derivation of the Fisher discriminant can be extended to find a subspace which appears to contain all of the class variability. LDA is based upon the concept of searching for a linear combination of variables (predictors) that best separates two classes (targets). The LDA algorithm is suggested by [15, 22]. To capture the notion of reparability, Fisher defined the following score function.

$$z = \beta_1 x_1 + \beta_2 x_2 + \cdots + \beta_d x_d \text{ ; Score function } s(\beta) = \frac{(\beta'\mu_1 - \beta'\mu_2)}{\beta'C\beta} ;$$

$$S(\beta) = \frac{(\overline{z_1} - \overline{z_2})}{Variance \text{ } of \text{ } z \text{ } within \text{ } groups}$$

Given the score function, the problem is to estimate the linear coefficients that maximize the score which can be solved by the following equations.

$$\beta = C^{-1}(\mu_1 - \mu_2); C = \frac{1}{n_1 + n_2}(n_1 C_1 + n_2 C_2)$$

One way of assessing the effectiveness of the discrimination is to calculate the Mahalanobis distance between two groups. A distance greater than 3 means that in two averages differ by more than 3 standard deviations. It means that the overlap (probability of misclassification) is quite small.

$$\Delta^2 = \beta'(\mu_1 - \mu_2)$$

Finally, a new point is classified by projecting it onto the maximally separating direction and classifying it as $C \text{ } I$ if: $\beta' \left[ X - \left( \frac{\mu_1 + \mu_2}{2} \right) \right] > -\log \frac{p(c_1)}{p(c_2)}$.

### 2.1.6. k-nearest neighbors

In pattern recognition, the *k*-nearest neighbors algorithm (*k*-NN) is a non-parametric method used for classification and regression. *k*-NN is a form of <u>instance-based learning</u>, or <u>lazy learning</u>, where the function is only approximated locally and all computation is deferred until classification. The *k*-NN algorithm is among the simplest of all <u>machine learning</u> algorithms. *k*-NN has some strong <u>consistency</u> results. The algorithm of K-NN is described by [4, 23]. There are other ways of calculating distance, and one way might be preferable depending on the problem we are solving. However, the straight-line distance (also called the Euclidean distance) is a popular and familiar choice. The KNN Algorithm-
 1. Select the dataset
 2. Set K to your selected number of neighbors
 3. Compute the distance between the query case and the current case from the data; Add the distance and the index of the case to an ordered group

4. Category the ordered group of distances and indices from smallest to largest (in ascending order) by the distances
   5. Choice the first K entries from the arranged group
   6. Catch the labels of the designated K entries
   7. If regression, return the mean of the K labels
   8. If classification, return the mode of the K labels.

2.1.7. Logistic Regression

Logistic regression is fundamentally a linear model for classification rather than regression. In this model, we use logistic regression to model probabilistically designated results of a single trial. It is a simple model which labels dichotomous output variables and can be extended for disease classification prediction. The algorithm of logistic regression is derived by [15, 22]. Once a logistic regression function has been recognized, and using training sets for each of the two populations, we can proceed to classify. Priors and costs are hard to combine into the analysis, so the classification rule develops, allocate z to population 1 if the estimated odds ratio is greater than 1 or

$$\frac{p(z)}{1-p(z)} = \exp(\beta_0 + \beta_1 z_1 + \cdots + \beta_r z_r) > 1$$

Consistently, we have the simple linear discriminant rule, allocate z to population 1 if the linear discriminant is greater than 0 or estimated odds ratio is greater than 1 or

$$\ln \frac{p(z)}{1-p(z)} \approx (\beta_0 + \beta_1 z_1 + \cdots + \beta_r z_r) > 0$$

2.1.8. Naive Bayes classification

In machine learning, naive Bayes classifiers are a family of simple "probabilistic classifiers" based on applying Bayes' theorem with strong (naive) independence assumptions between the features.

The associations between dependent events can be labelled using Bayes' theorem, as shown in [13, 21, 29]. Given a set of variables, $X = \{x_1, x_2, \cdots, x_d\}$, we want to build the posterior probability for the event $C_j$ among a set of potential results $C = \{c_1, c_2, \cdots, c_d\}$. In a more familiar language, X is the predictors and C is the set of categorical levels present in the dependent variable. Using Bayes' rule: $p(C_j \mid x_1, x_2, \cdots, x_d) \propto p(C_j \mid x_1, x_2, \cdots, x_d) p(C_j)$ where $p(C_j \mid x_1, x_2, \cdots, x_d)$ is the posterior probability of class membership, i.e., the probability that X belongs to $C_j$. Since Naive Bayes assumes that the conditional probabilities of the independent variables are statistically independent we can decompose the likelihood to a product of terms: $p(X \mid C_j) \propto \prod_{k=1}^{d} p(x_k \mid C_j)$ and rewrite the posterior as:

$$p(X \mid C_j) \propto p(C_j) \prod_{k=1}^{d} p(x_k \mid C_j)$$

Using Bayes' rule above, we label a new case X with a class level $C_j$ that achieves the highest posterior probability.

2.2. *Performance Evaluation Criteria for Algorithm*

This section presents measure for assessing how good or how accurate our classifier is at predicting the class level of tuples described by [15, 26]. The entries in the confusion matrix have the following meaning in the context of our study. $T_p$ is the number of correct predictions that an instance is positive. $F_n$ is the number of incorrect predictions that an instance is negative. $F_p$ is the number of incorrect predictions that an instance is positive and $T_n$ is the number of correct predictions that an instance is negative. These terms are summarized in the confusion matrix of Table 1.

**Table 1.** Confusion matrix

|        |          | Predicted |          |
|--------|----------|-----------|----------|
|        |          | Positive  | Negative |
| Actual | Positive | $T_p$     | $F_n$    |
|        | Negative | $F_p$     | $T_n$    |

The classifier evaluation measures presented in this section are summarized in Table 2.

**Table 2**. Classifier evaluation measures

| Tools | Statistic |
|---|---|
| **Recall /Sensitivity** | $R = T_p / (T_p + F_n)$ |
| Precision | $P = T_p / (T_p + F_p)$ |
| Accuracy | $A = (T_p + T_n) / (T_p + T_n + F_p + F_n)$ |
| F-Score | $F = (2 \times \mathrm{Precision} \times \mathrm{Recall}) / (\mathrm{Precision} + \mathrm{Recall})$ |

*2.3. Software used*

The Machine learning classification algorithms in this research are employed based on R 3.2.4 version is open source software written in [30], an open gathering of machine learning algorithms tolerates the researcher to excavation his own data for fashions and designs. The algorithms can either be applied straight to a dataset or called from the investigator own R code. R holds tools for data pre-processing, classification, regression, clustering, association rules, and visualization. In our experiment we performed different supervised ML classification algorithms.

3. Results

The investigation has been carried out using a small educational example taken from marine biology. These are biological and environmental observations made at 30 sampling points on the sea bed. The number of 30 in this case is more realistic because there are few sampling locations in marine environmental sampling. The data have five variable, three measurements and one classification such as five species, labelled a-e; the values of depth x (in meter), pollution index y, the temperature z in 0C and the sediment type (three categories). The classification methods that we used here –Decision Tree, Random Forests, Artificial Neural Network, Support Vector Machines, Linear Discrimination Analysis, k-nearest neighbors, Logistic Regression and Naive Bayes. Firstly we applied some basic statistical tools to visualize the data then we applied afore-mentioned algorithms in the same data set. The outcomes of the mentioned methods are given in Figure 2-6.

From Figure 2, Pairwise Scatterplots of the eight variables showed in each case the smooth relationship of the vertical variable with respect to the horizontal one, the lower triangle gave the correlation coefficients, with size of numbers proportional to their absolute value. Representation of ecological data set in a three dimensional space. The figure appear to be three groups, which we have represented using different colors. In Figure 3 we showed the relationship between the depth, pollution and temperature settings used to separate the class. Result from heat map and parallel coordinate in Figure 4 & 5 respectively perform on the ecological data set. The two figures are not clearly separate the data set and so these figures are fail to identify the actual class sizes. Each panel of Figure 6 is a diagram for a variables whose identities are given by the corresponding scatter plots, densities, correlations and box plots. A classification data set involving three groups. Each group is shown using different colored symbol. The three groups are almost separated. In this setting, a classification approach is successfully identify the three classes.

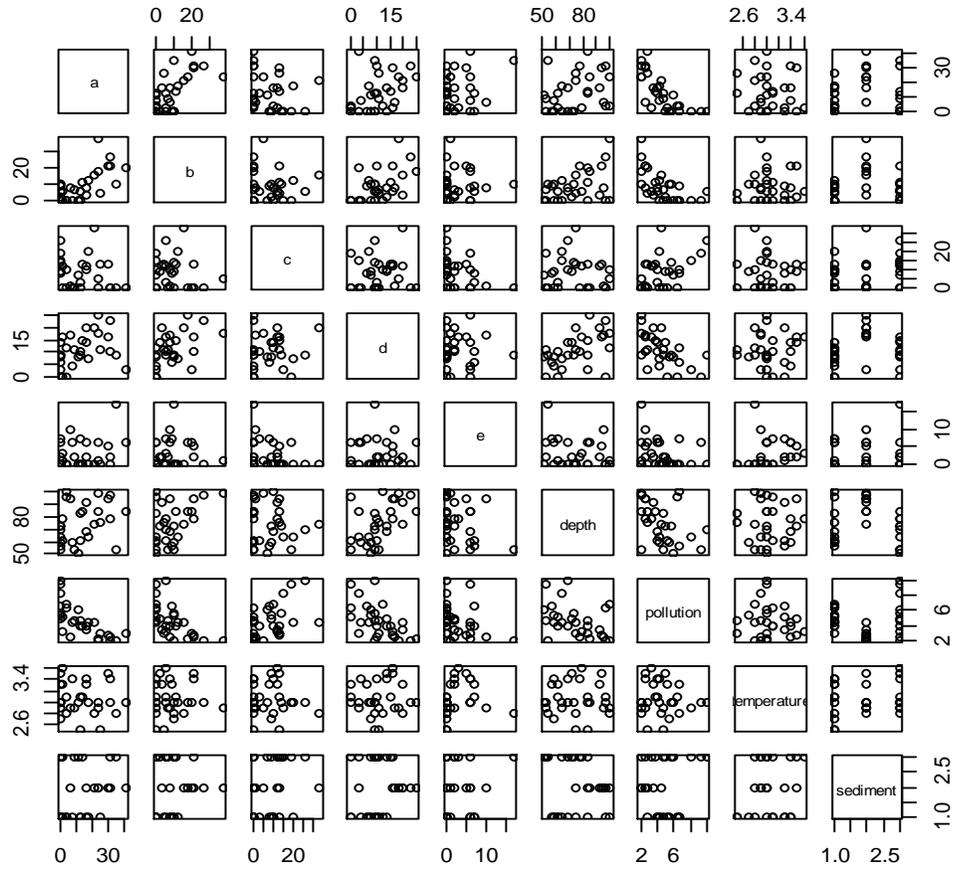

**Figure 2**. A Matrix of Scatter Plots

**Figure 3.** 3D Scatter Plot

**Figure 4.** Heat Map

**Figure 5.** Parallel Coordinates

**Figure 6.** Information about scatter plots, densities, correlations and box plots

The classification tree that results from using decision tree methodology with the ecological data is shown in Figure 7 & 8. The splitting rules are indicated in Figures. It is clear that the data set contain three classes. We see that the variables do a very good job of distinguishing the three classification. Moreover, error rate of random forest are shown in Figure 9 for the ecological data set. It is clean from the figure that the ecological variables separate the three groups quite well. Table 3 presents mean decrease Gini. Figure 10 & 11 have a supervising characteristics. A two dimensional classification figures in which the tree decision boundary is linear. Here, the model is unable to separate the classification, although it is not clear where as a decision tree is successful. The network diagram for the 205 steps neural network. The input, hidden, and the output variables are represented by nodes, and the weight parameters are represented by links between the nodes. Which is notify in Figure 12. Table 4 shows the result of SVM classification, it was observed that the algorithm revealed three classes in the dataset Table 5 shows the LDA results. The separation of the three groups is fully exhibited in the two dimensional discriminant space. The scatter of the observations in the discriminant coordinate system are shown in Figure 13. The separation is quite good. The plot in Figure 14 indicates that LD1 and LD2 are the best choice for the dimension of the final configuration. A representation of the ecological data set followed by partition plot is perfectly classify that indicated the three classes. A K-near neighbors (KNN) algorithm applied to the ecological data gives the two dimensional representation shown in Figure 15 & 16. Both figures representation are very similar. It is clear that the graphs are almost three classes. Using the K-NN algorithm, the software choose K values and the exhibited the training error and test error which are very close to each other. It is clear that misclassification error is very low here. A plot of error for the ecological variables is shown in Figure 17. A plot of prediction using logistic regression is shown in Figure 18. Here the separation on the basis of the prediction values is not good. Figure 19, which shows the three normal densities for the classification explains this phenomenon. Use of the midpoint among the three densities does not make the three misclassification probabilities equal. It is clearly penalizes the population with the three classification.

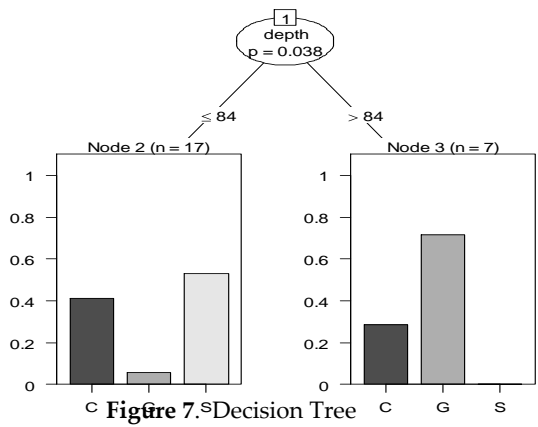

**Figure 7.** Decision Tree

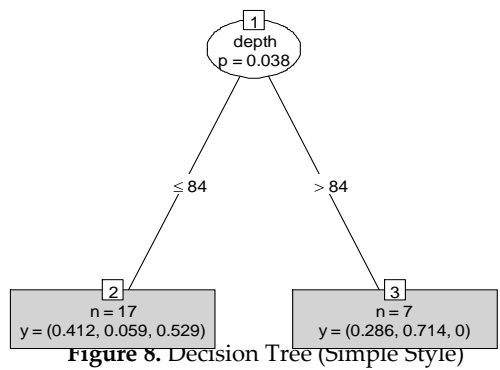

**Figure 8.** Decision Tree (Simple Style)

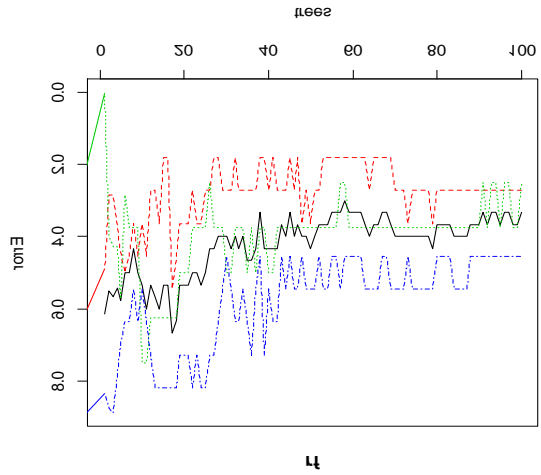

**Figure 9**. Error Rate of Random Forest

| Table 3. Mean decrease Gini | |
|---|---|
| Variable | Mean Decrease Gini |
| a | 2.4128953 |
| b | 2.9298969 |
| c | 1.7157882 |
| d | 3.4177267 |
| e | 0.8432125 |
| Depth | 3.5373955 |
| Pollution | 2.4515585 |
| Temperature | 1.8348597 |

Mean Decrease Gini represents the mean decrease in node impurity (and not the mean decrease in accuracy).

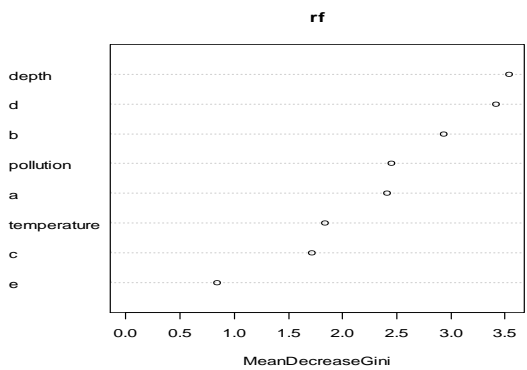

**Figure 10**. Variable importance

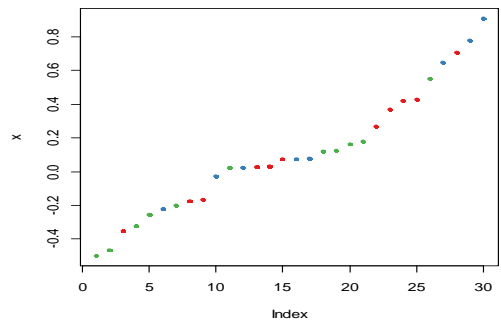

**Figure 11**. Margin of predictions

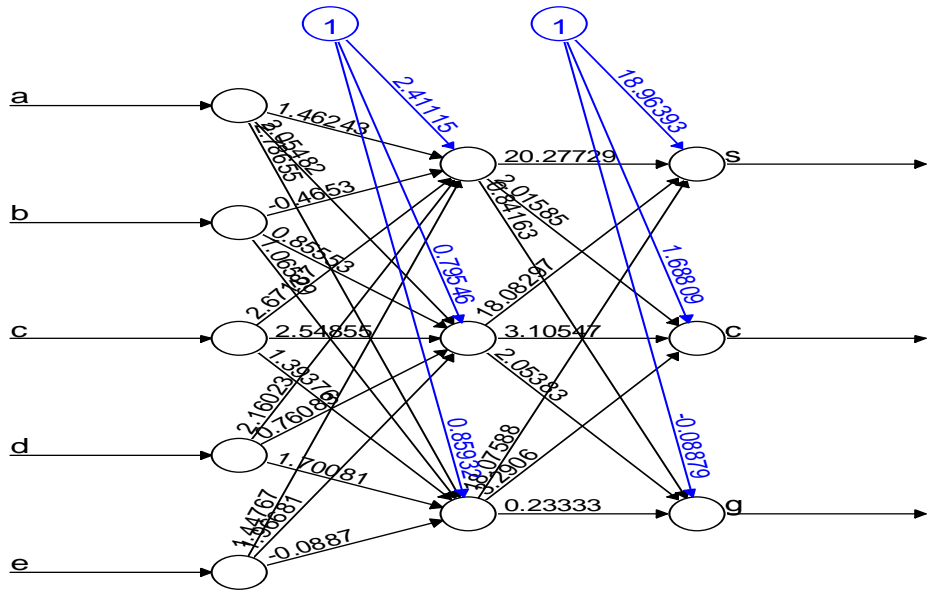

Error: 1663.992   Steps: 205

**Figure 12**. Neural net for 3 class problem

**Table 4.** Parameters of SVM classification

| SVM-Type | C-classification |
|---|---|
| SVM-Kernel | radial |
| Cost | 1 |
| Gamma | .125 |
| Number of Support Vectors | 30 |
| Number of Classes | 3 |
| Levels | C, G, S |

**Table 5**. Coefficients of linear discriminant

| Variable | LD1 | LD2 |
|---|---|---|
| a | 0.03131988 | 0.03391143 |
| b | 0.06775184 | 0.04446228 |
| c | 0.01531784 | -0.0139997 |
| d | 0.10400125 | 0.14842507 |
| e | 0.11503986 | 0.03094287 |
| Depth | 0.04686593 | -0.0479481 |
| Pollution | 0.18435519 | 0.63597867 |
| Temperature | -0.04162924 | 1.68329544 |

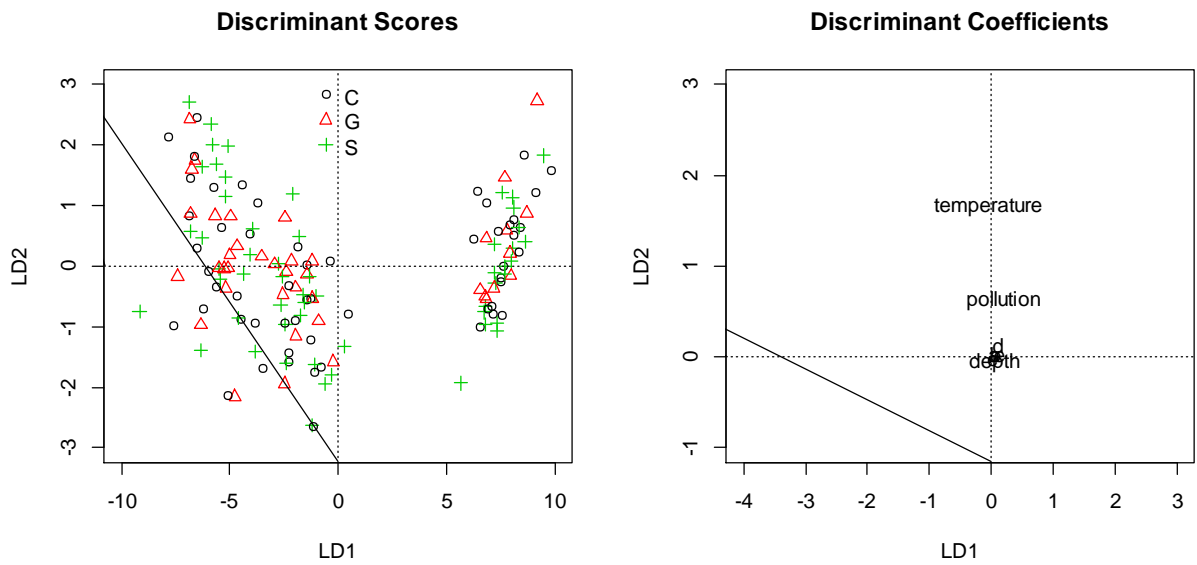

**Figure 13**. Discriminant score and coefficient

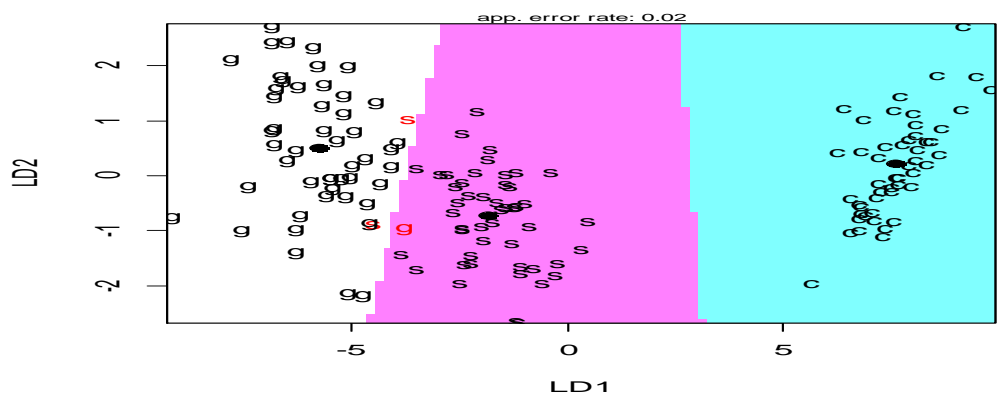

**Figure 14.** Partition plot

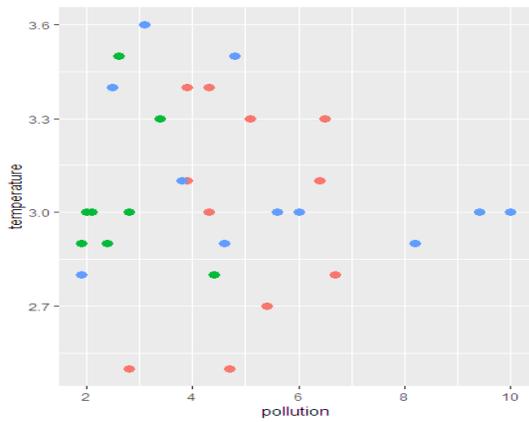

**Figure 15**. temperature versus pollution by Sediment

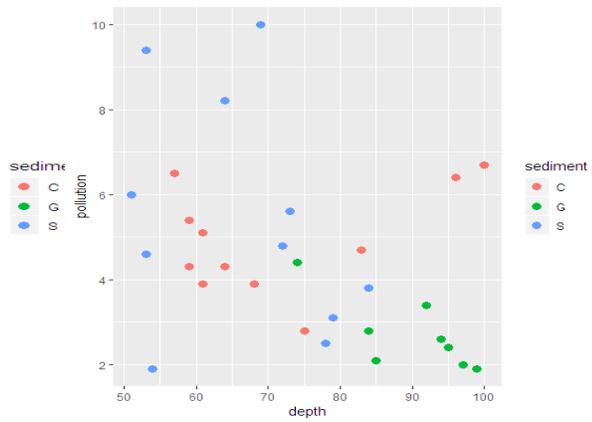

**Figure16.** Depth versus pollution by sediment

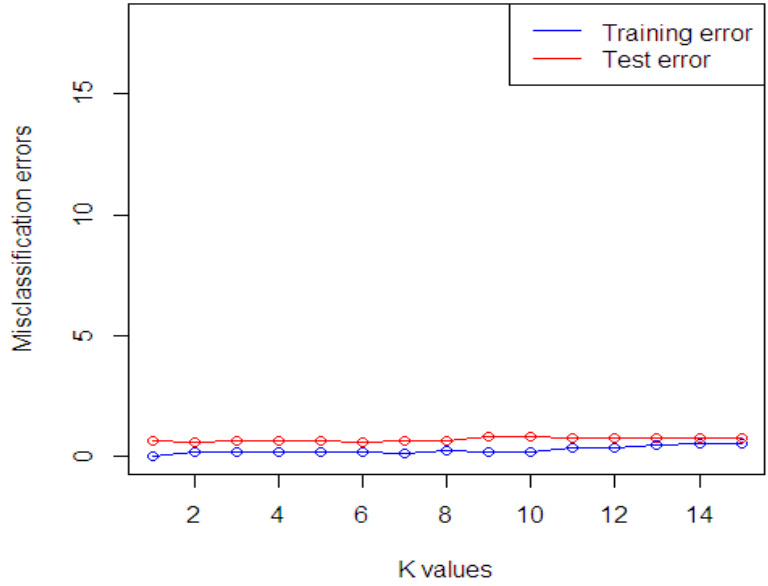

**Figure 17.** Training and test accuracy using KNN with different K's

| Data control | | Predicted Value |
|---|---|---|
| Pollution | Sediment | |
| 2.8 | C | 0.6598954 |
| 8.2 | S | 0.4648191 |
| 6.5 | C | 0.5279924 |
| 3.8 | S | 0.6257388 |
| 9.4 | S | 0.4207793 |
| 4.7 | C | 0.5938767 |
| 6.7 | C | 0.5205677 |
| 2.8 | G | 0.6598954 |
| 3.1 | S | 0.6498030 |
| 4.3 | C | 0.6081529 |

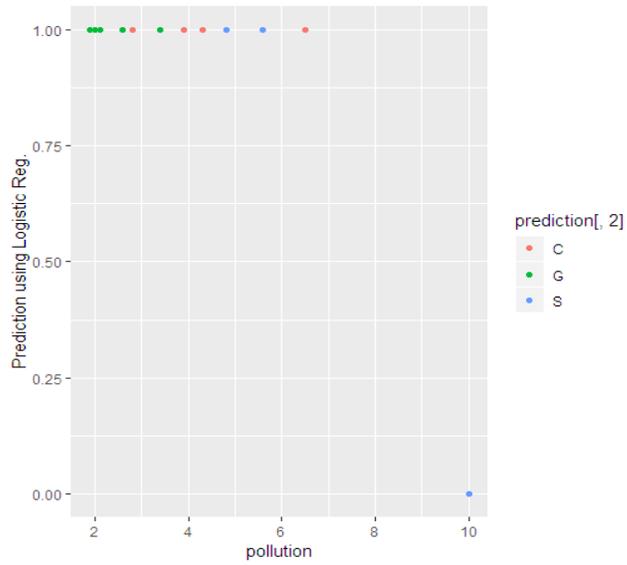

**Table 6.** Predicted value  **Figure 18.** Prediction Plot

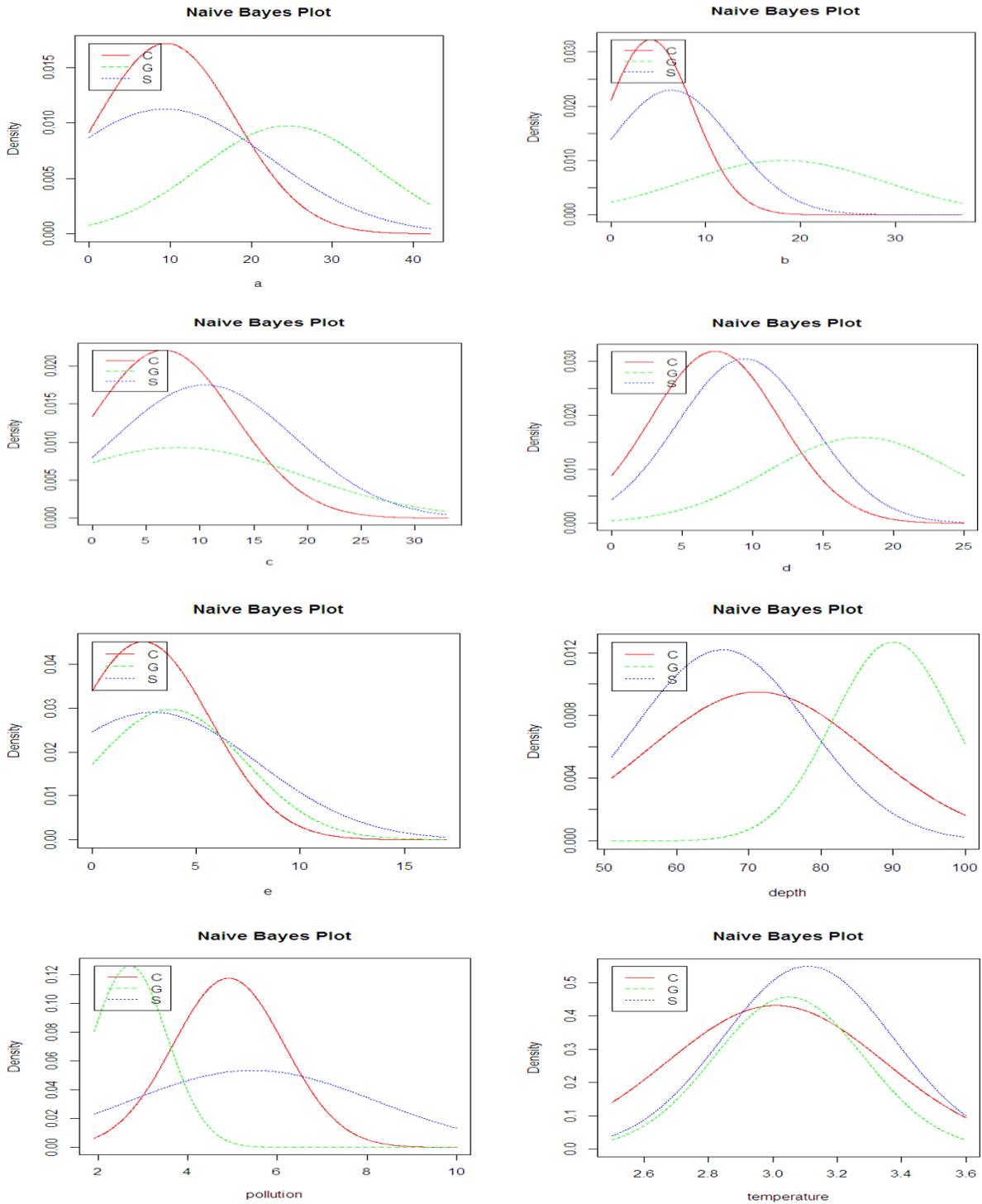

**Figure 19.** Naive base classification

**Performance Evaluation Process**

This research paper investigates and compares applicability of different types of supervised ML classifiers to predict the ecological data set. It is necessary in our research to apply the following processes for analyzing data set:

(i). Process I

The research is conducted by using the whole data set as testing data set and the results are illustrated in Table 7.

(ii). Process II

The research is conducted based on percentage split of the whole data into seventy five percentage of the data for training the classification model and twenty five percentage of the data for testing the model as illustrated in Table 8.

(iii). Process III

The research is conducted by using 3 Fold cross validation as testing option and the results are shown in Table 9.

Table 7 shows the classification accuracy and other performance measures results from conducting process I of applying supervised ML algorithms and the results as follows: LDA and NB classifiers are superior in their accuracy results where it could classify above 95% all the instances correctly, and so they have the highest measures in ecological data set. Alternatively ANN classifier has lowest accuracy since it classify correctly only 78% from the whole data. Table 8 shows the supervised ML classification accuracy and other performance measures results from conducting process II in which the whole data set is divided into two splits; 75% for training the classifiers, and 25% for testing the classifier of applying the different types of classifiers and perform a comparative analysis as follow: NB and LDA have the highest classification accuracy among the other classifiers. Alternatively, K-NN showed the worst result. The accuracy results and overall performance measures are explained in Table 9 as follow: LDA and NB algorithms have explored the highest accuracy as well as highest recall/sensitivity, precision, accuracy and F-score in this process. But K-NN performs badly; it could not classify correctly the data. The Comparison between various classifications algorithms on given datasets are represented in Figure 20.

Table 7. Performance of process I

| Algorithm | $T_p$ | $F_p$ | $T_n$ | $F_n$ | Recall / Sensitivity | Precision | Accuracy | F-Score |
|---|---|---|---|---|---|---|---|---|
| DT | 0.9151 | 0.0848 | 0.7615 | 0.2384 | 0.7932 | 0.91519 | 0.83836 | 0.8498 |
| RF | 0.9887 | 0.0112 | 0.8976 | 0.1023 | 0.9061 | 0.98873 | 0.94317 | 0.9456 |
| ANN | 0.8653 | 0.1346 | 0.7131 | 0.2868 | 0.7510 | 0.86532 | **0.789235** | 0.8041 |
| SVM | 0.8966 | 0.1033 | 0.7866 | 0.2134 | 0.8077 | 0.89663 | 0.841615 | 0.8498 |
| LDA | 0.9987 | 0.0012 | 0.9325 | 0.0675 | **0.9366** | **0.9987** | **0.9656** | **0.9667** |
| K-NN | 0.8913 | 0.1086 | 0.7231 | 0.2769 | 0.7629 | 0.89134 | 0.80722 | 0.8221 |
| LR | 0.9028 | 0.0971 | 0.7718 | 0.2281 | 0.7982 | 0.90287 | 0.83737 | 0.8473 |
| NB | 0.9907 | 0.0092 | 0.9123 | 0.087 | **0.9186** | **0.99074** | **0.95152** | **0.9533** |

Table 8. Performance of process II

| Algorithm | $T_p$ | $F_p$ | $T_n$ | $F_n$ | Recall / Sensitivity | Precision | Accuracy | F-Score |
|---|---|---|---|---|---|---|---|---|
| DT | 0.6573 | 0.3427 | 0.5631 | 0.4369 | 0.6007 | 0.6573 | 0.6102 | 0.6277 |
| RF | 0.7018 | 0.2982 | 0.6983 | 0.3017 | 0.6993 | 0.7018 | 0.7000 | 0.7005 |
| ANN | 0.6869 | 0.3131 | 0.7838 | 0.2162 | 0.7606 | 0.6869 | 0.7353 | 0.7218 |
| SVM | 0.5653 | 0.4347 | 0.6184 | 0.3816 | 0.5970 | 0.5653 | 0.5918 | 0.5807 |
| LDA | 0.8168 | 0.1832 | 0.7978 | 0.2022 | **0.8015** | **0.8168** | **0.8073** | **0.8091** |
| K-NN | 0.5931 | 0.4069 | 0.5632 | 0.4368 | 0.5758 | 0.5931 | **0.5781** | 0.5843 |
| LR | 0.6751 | 0.3249 | 0.6087 | 0.3913 | 0.6330 | 0.6751 | 0.6419 | 0.6534 |
| NB | 0.8332 | 0.1668 | 0.7996 | 0.2004 | **0.8061** | **0.8332** | **0.8164** | **0.8194** |

Table 9. Performance of process III

| Algorithm | $T_p$ | $F_p$ | $T_n$ | $F_n$ | Recall / Sensitivity | Precision | Accuracy | F-Score |
|---|---|---|---|---|---|---|---|---|
| DT | 0.6375 | 0.3625 | 0.7256 | 0.2744 | 0.6990 | 0.6375 | 0.6815 | 0.6668 |
| RF | 0.6889 | 0.3111 | 0.7489 | 0.2511 | 0.7328 | 0.6889 | 0.7189 | 0.7102 |
| ANN | 0.8573 | 0.1427 | 0.7956 | 0.2044 | 0.8074 | 0.8573 | 0.8264 | 0.8316 |
| SVM | 0.8254 | 0.1746 | 0.8397 | 0.1603 | 0.8373 | 0.8254 | 0.8325 | 0.8313 |
| LDA | 0.9137 | 0.0863 | 0.9835 | 0.0165 | **0.9822** | **0.9137** | **0.9486** | **0.9467** |
| K-NN | 0.6478 | 0.3522 | 0.6959 | 0.3041 | 0.6805 | 0.6478 | **0.6718** | 0.6637 |
| LR | 0.7756 | 0.2244 | 0.7466 | 0.2534 | 0.7537 | 0.7756 | 0.7611 | 0.7645 |
| NB | 0.9438 | 0.0562 | 0.9478 | 0.0522 | **0.9475** | **0.9438** | **0.9458** | **0.9456** |

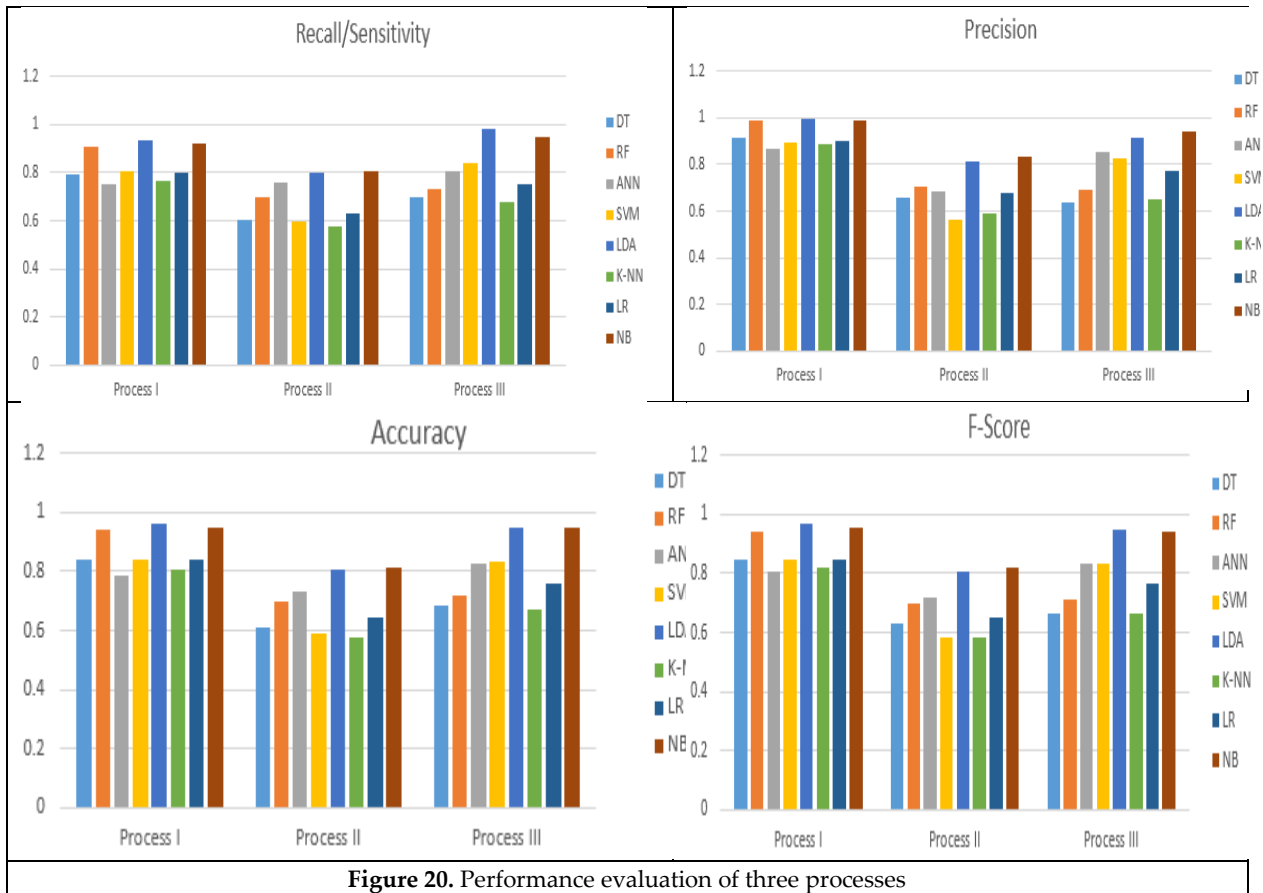

Figure 20. Performance evaluation of three processes

## 4. Conclusion

In this paper, we have inspected the execution of eight machine learning algorithms namely Decision Trees, Random Forest, Neural Network, Support Vector Machine, Linear Discriminant Analysis, k-nearest neighbors, Logistic Regression and Naive Bayes to correctly classify the ecological data. The utility of the methods is further demonstrated in terms of recall/sensitivity, precision, accuracy and F-score obtained from three processes. The overall results of the conducted data set showed that LDA & NB classification algorithms are considered an accurate classifiers and outperformed many other supervised ML classification algorithms in various situation. Further, we plan to re-form our study of Classification models by introducing the intelligent machine learning algorithms useful to a large collection of real life data set.

**Author Contributions:** Md.Siraj-Ud-Doulah has contributed this paper for his constant support of preparing methodology, formal analysis based on real life data set and check the validity by using various performance measures, investigation followed by software R programming. Alternatively, Md.Ashed Alam has contributed in this paper for his continuous care of supervision based on visualization of the task. Finally, in this work, draft preparation, review and edit have been made by both of them.

**Funding:** This research received no external funding.

**Conflicts of Interest:** The authors declare no conflict of interest.

**References**

1. Alam M.A, Lin H. Y. Deng H.W. Calhoun, V.D and Wang, Y.P. A kernel machine mehod for detecting higher order interactions in multimodal datasets: Application to schizophrenia, Journal of neuroscience methods, 2018; 309, 161-174.
2. Alam M.A, Calhoun, V.D and Wang, Y.P. Identify outliers using multiple kernel canonical correlation analysis with application to imaging genetis. Computational Statistics and Data Analysis, 2018; 125, 70-85.
3. Alam M. A and Fukumizu K. Higher-order regularized canonical correlation analysis, International Journal of Pattern Recognition and Artificial Intelligence, 2015; 29(4), 1-24.
4. Altman, N. S. An introduction to kernel and nearest-neighbor nonparametric regression. The American Statistician. 1992; 46 (3), 175–185.
5. Ali, J., Khan, R,, Ahmad, N., Maqsood,I. Random forests and decision tree, Journal of Computer Science, 2012; 9(5),272-278.
6. Alex, S., Vishwanathan, S.V.N. Introduction to Machine Learning. 1st ed.; University of Cambridge, Cambridge, United Kingdom, 2008; 181-186.
7. Brazdil, P., Giraud-Carrier, C., Soares, C. Meta learning: Applications to Data Mining, 1st ed.; Springer Verlag, Berlin Heidelberg, 2009; 17-42.
8. Ben, H., Asa, Horn, David, Siegelmann, Hava, Vapnik, Vladimir N. Support vector clustering, Journal of Machine Learning Research 2001; 2, 125–137.
9. Cherkassky, V., Mulier, F. M. Learning from Data: Concepts, Theory,and Methods, 2nd ed.; John Wiley- IEEE Press, 2007; 340-464.
10. Glenn D. Multivariate Regression Trees: A New Technique for Modeling Species–Environment Relationships. Ecological Society of America, 2002; 83 (4), 1105-1117.
11. Guobin, Z., Dan G. B. Classification using ASTER data and SVM algorithms: The case study of Beer Sheva, Israel. Remote Sensing of Environment, 2002; 80(2), 233-240.
12. Greenacre, M. & Promicerio, R. Multivariate Analysis of Ecological Data, 1st ed.; Rubes Editorial, Spain, 2013; 15-24.
13. Goyal,A., Mehta,R Performance Comparison of Naive Bayes and J48 Classification Algorithms, IJAER, 2012; 7(11), 281-297.
14. Hormozi, H., Hormozi, E. & Nohooji, H. R. The Classification of the Applicable Machine Learning Methods in Robot Manipulators. International Journal of Machine Learning and Computing (IJMLC), 2012; 2(5), 560-563.
15. Han, J, Kanber, M. Pei J. Data Mining: Concepts and Techniques, 3rd ed.; Morgan Kaufman, USA, 2012; 327-439.
16. Landwehr, N., Hall, M., Frank, E. Logistic model trees, Machine Learning, 2005; 59:161-205.
17. Ilyes, J., Nahla, B. A., Zied, E. Overview of Decision Trees as Possibilistic Classifiers. International Journal of Approximate Reasoning, 2008; 48:784–807.
18. Kuncheva, L. Combining pattern classifiers: Methods and algorithms, John Wiley & Sons, 1st ed.; West Sussex, England, 2004; 45-99.

19. Kellie, J, Ryan, V.K. Empirical characterization of random forest variable importance measures, Journal of Computational Statistics & Data Analysis, 2008; 52, 2249-2260.
20. Kotsiantis, S. B., Zaharakis, I. D., Pintelas, P. E. Machine learning: a review of classification and combining techniques, Artif Intell Rev, 2007; 26:159–190.
21. Margaret, H., Danham, S., Sridhar. Data mining, Introductory and Advanced Topics, Person education, 1st ed.; UK, 2006; 75-84.
22. McLachlan, G. J. Discriminant Analysis and Statistical Pattern Recognition. Wiley Interscience, 1st ed.; UK, 2004; 189-200.
23. Nigsch, F, Bender, A, Buuren, B, Tissen, J, Nigsch, E, Mitchell, J.B. Melting point prediction employing k-nearest neighbor algorithms and genetic parameter optimization. Journal of Chemical Information and Modeling. 2006; 46 (6), 2412–2422.
24. Osisanwo F.Y, Akinsola J.E.T., Awodele O., Hinmikaiye J. O., Olakanmi O., Akinjobi J. Supervised Machine Learning Algorithms: Classification and Comparison, International Journal of Computer Trends and Technology (IJCTT),2017; 48( 3),128-138.
25. Pradeep, K. R. & Naveen, N. C. A Collective Study of Machine Learning (ML) Algorithms with Big Data Analytics (BDA) for Healthcare Analytics (HcA). International Journal of Computer Trends and Technology (IJCTT), 2017; 47(3), 149 – 155.
26. Lior, R., Maimon, O. Data mining with decision trees: theory and applications. 1st ed.; World Scientific Pub Co. Inc., USA, 2008; 31-58.
27. Shujun, H., Nianguang, C., Pedro, P., Pacheco, Shavira, N., Yang, W., Wayne, X. Overview of Applications of Support Vector Machine (SVM) Learning in Cancer Genomics. Cancer Genomics Proteomics. 2018; 15(1), 41–51.
28. Sharma, A.K., Sahni, S. A Comparative Study of Classification Algorithms for Spam Email Data Analysis, IJCSE, 2011; 3(5), 1890-1895.
29. Witten, I. H. and Frank, E. Data Mining: Practical Machine Learning Tools and Techniques, 3rd ed.; Morgan Kaufmann, USA, 2011; 203-215.
30. Crawley, M.J. The R Book, 1st ed.; John Wiley & Sons Ltd, England, 2007; 811-827.